\documentclass[runningheads]{llncs}

\usepackage{booktabs} % add this in preamble
\usepackage[T1]{fontenc}
\usepackage{graphicx} % Required for inserting images
\usepackage{float}
\usepackage{hyperref} %for using itemize
\usepackage{tikz} %for drawing image and axis
\usepackage{dsfont}
\usetikzlibrary{arrows.meta, positioning, shapes.geometric, calc}
\usepackage{amsmath, amssymb, amsfonts} % for math
\usepackage{booktabs} %for table
\usepackage{textcomp}
\usepackage[numbers]{natbib}

\usepackage{xcolor}
\usepackage{caption} % for \captionsetup
\usepackage{algorithmic}
\usepackage{tabularx}
\usepackage{pgfplots} %for drawing axis
\pgfplotsset{compat=1.18}
\usepackage{orcidlink}
\begin{document}

\title{MiRAGE: Misconception Detection with Retrieval-Guided Multi-Stage Reasoning and Ensemble Fusion}

\author{
Van Duc Cuong\inst{1}\orcidID{0009-0008-6700-1477} \and
Tran Quoc Thai\inst{1}\orcidID{0009-0000-7129-4833} \and
Nguyen Dinh Tuan Minh\inst{1}\orcidID{0009-0003-9141-7409} \and
Vu Duc Tam\inst{1}\orcidID{0009-0006-4506-531X} \and
Nguyen Van Son\inst{2}\orcidID{0000-0001-8188-4984} \and
Nguyen Thi Hanh\inst{3,*}\orcidID{0009-0006-7929-3329}}

\authorrunning{Cuong et al.}

\titlerunning{MiRAGE}

\institute{Hanoi University of Science and Technology, Hanoi, Vietnam.  \and School of Computing, Phenikaa University, Hanoi, Vietnam. \and
Faculty of Interdisciplinary Digital Technology, Phenikaa University, Hanoi, Vietnam.
\\ \inst{*}Corresponding author at Phenikaa University, Hanoi, Vietnam\\
% \email{\{25900004\}@st.phenikaa-uni.edu.vn}\\
\email{\{cuong.vd220021, thai.tq230065, minh.ndt230048, tam.vd230064\}@sis.hust.edu.vn}\\
\email{\{son.nguyenvan, hanh.nguyenthi\}@phenikaa-uni.edu.vn}
}

\maketitle
\begin{abstract}
% Detecting student misconceptions remains a longstanding challenge in education, demanding fine-grained semantic understanding and logical reasoning over open-ended responses. In this paper, we introduce \textsc{MiRAGE}—Chain-of-Thought Reasoning and Ensemble Framework—a cost-effective hybrid approach for misconception detection in mathematics. \textsc{MiRAGE} integrates three key modules: (1) a \emph{Retrieval module} that selects semantically related candidates from the dataset; (2) a \emph{Reasoning module} that applies Chain-of-Thought reasoning to uncover logical inconsistencies in student solutions; and (3) a \emph{Reranking module} that leverages cross-attention scoring to refine predictions. These components are unified through an ensemble fusion mechanism, providing both interpretability and robustness. Experimental results show that \textsc{MiRAGE} achieves Mean Average Precision scores of 0.82/0.92/0.93 at levels 1/3/5, outperforming individual modules and demonstrating the strength of the ensemble strategy. Additional findings highlight the importance of fine-tuned reasoning and reranking models in improving overall performance. By reducing reliance on large-scale large language models while maintaining high accuracy, \textsc{MiRAGE} delivers a scalable and practical solution for automated misconception detection in mathematics education.

Detecting student misconceptions in open-ended responses is a longstanding challenge, demanding semantic precision and logical reasoning. We propose \textsc{MiRAGE} - \textit{Misconception Detection with Retrieval-Guided Multi-Stage Reasoning and Ensemble Fusion}, a novel framework for automated misconception detection in mathematics. \textsc{MiRAGE} operates in three stages: (1) a \emph{Retrieval module} narrows a large candidate pool to a semantically relevant subset; (2) a \emph{Reasoning module} employs chain-of-thought generation to expose logical inconsistencies in student solutions; and (3) a \emph{Reranking module} refines predictions by aligning them with the reasoning. These components are unified through an ensemble-fusion strategy that enhances robustness and interpretability. On mathematics datasets, \textsc{MiRAGE} achieves Mean Average Precision scores of 0.82/0.92/0.93 at levels 1/3/5, consistently outperforming individual modules. By coupling retrieval guidance with multi-stage reasoning, \textsc{MiRAGE} reduces dependence on large-scale language models while delivering a scalable and effective solution for educational assessment.

\end{abstract}

\section{Introduction}

Understanding how learners think—and where their reasoning goes astray—remains a longstanding challenge in education. As students engage with new material, they draw upon prior knowledge, intuition, and individual reasoning strategies that shape their interpretation of concepts. While these cognitive processes are central to learning, they can also lead to systematic misunderstandings, or misconceptions, that persist over time. Identifying such misconceptions is critical for guiding instruction and improving educational outcomes, yet achieving this reliably and at scale remains a significant challenge. This motivates the development of automatic systems for misconception detection, which can reduce the cost and effort required from teachers and provide students with timely, personalized support in their self-directed learning.

Parallel to this motivation, the growing availability of student-generated data, particularly open-ended responses, creates an opportunity to leverage recent advances in artificial intelligence to capture student reasoning better. Specifically, Natural language processing (NLP) models \cite{fanni2023natural} offer a promising approach for analyzing such responses and identifying potential misconceptions. In the current real-world context, popular NLP models—huge language models (LLMs) \cite{zhao2023survey} such as GPT \cite{achiam2023gpt}, Qwen \cite{yang2025qwen3}, and Gemma \cite{team2025gemma}—have shown remarkable effectiveness across a broad spectrum of tasks. Nevertheless, while these LLMs are not pretrained with a dedicated focus on specialized domains of educational reasoning, their direct application also entails significant costs through commercial service fees or the substantial computational resources required for local deployment. Moreover, misconception detection in education introduces additional challenges, as it demands domain-specific semantic understanding and logical reasoning to uncover subtle errors in students’ explanations. These considerations motivate exploring whether smaller models can provide a more practical and cost-effective alternative when equipped with reasoning enhancements and knowledge distillation.

Addressing the aforementioned challenges, this work introduces \textbf{MiRAGE}, a hybrid framework that leverages the collaboration of small- to medium-sized language models (LMs), along with Chain-of-Thought (CoT) reasoning techniques to develop a cost-effective automatic system for the task of misconception detection. The targeted domain is mathematics, which inherently requires substantial logical reasoning. Through this approach, we aim to enable scalable and accurate identification of misconceptions, contributing to the broader objective of delivering personalized feedback and enhancing learning experiences.

\section{{Related works}}

Early works on misconception detection primarily relied on handcrafted rules or item response theory to analyze student errors. For example, traditional methods such as think-aloud protocols were used to identify errors in students' mathematical reasoning (\citet{secolsky2016thinkaloud} \cite{secolsky2016thinkaloud}). More recently, data-driven approaches have emerged that move beyond traditional psychometric models. Probabilistic models have been employed to analyze students' written explanations, while language models have shown promise in automatically recognizing and categorizing misconceptions from free-text responses (\citet{michalenko2017dataminingtextualresponsesuncover} \cite{michalenko2017dataminingtextualresponsesuncover}).

Beyond error detection, recent studies have leveraged CoT (\citet{wei2023chainofthoughtpromptingelicitsreasoning} \cite{wei2023chainofthoughtpromptingelicitsreasoning}) to enhance transparency and improve reasoning accuracy, as step-by-step explanations help LLMs perform better on reasoning and classification tasks (\citet{kostina2025llmTextClassification} \cite{kostina2025llmTextClassification}), thereby offering a more interpretable decision-making process.

In addition to these directions, prior studies have also explored techniques such as reranking for improving text matching in classification tasks (Setiawan et al.\cite{setiawan2024reranking}) and knowledge distillation for enhancing model efficiency (\citet{calderon2023systematicstudyknowledgedistillation} \cite{calderon2023systematicstudyknowledgedistillation}). Together, these approaches demonstrate the potential of reusing large pretrained models and aligning them with lightweight architectures that are particularly suitable for practical educational applications.

Ensemble learning, though less explored in misconception detection, has proven effective in educational NLP and text classification by improving robustness and accuracy. Integrating models such as transformers, graph neural networks, and probabilistic classifiers allows ensembles to balance strengths and weaknesses. Recent work shows that stacking diverse transformers enhances assessment accuracy \cite{huang2024improvingacademicskillsassessment}, while broader studies confirm that hybrid ensembles improve generalization in language tasks \cite{jia2024reviewhybridensembledeep, Ganaie_2022}.

Nevertheless, existing approaches face key limitations. Specifically, rule-based and probabilistic models generalize poorly, performing well on small datasets but failing to capture semantic variation in large-scale free-text responses. Meanwhile, LLMs raise concerns of transparency and reliability, as their lack of interpretability and tendency to hallucinate undermine trust. These challenges require more reliable, interpretable, and semantically adaptive ensemble frameworks that can balance accuracy with robustness.

To address these limitations, we propose MiRAGE, a hybrid framework that integrates CoT prompting with ensemble learning for misconception detection. Our main contributions are as follows:
\begin{itemize}
\item \textbf{Hybrid two-stage framework:} A pipeline combining similarity-based retrieval and a cross-attention reranker, conditioned on question, student answer, and reasoning, for more accurate classification.
\item \textbf{CoT integration:} Unlike prior reranking methods, we use CoT reasoning to generate intermediate explanations, improving interpretability and guiding the overall classification process.

\item \textbf{Verification-based reranking:} Reformulating reranking as a verification task with a logit-difference scheme, providing a simple yet effective objective.
\item \textbf{Knowledge distillation:} Transferring knowledge from LLMs into smaller models to retain performance while reducing inference cost, enabling large-scale applications.
\end{itemize}

%\begin{itemize}
%    \item \textbf{Hybrid two-stage framework:} We propose a novel pipeline that integrates a similarity-based retrieval process with a cross-attention reranker, conditioned on the question, student answer, explanation, and associated reasoning, to produce more accurate and robust classification decisions.

%    \item \textbf{CoT integration:} Unlike prior reranking approaches, our method leverages CoT reasoning to generate intermediate explanations, enhancing interpretability and guiding the final classification.
    
%    \item \textbf{Reranking with a verification objective:} Reranking is reformulated as a verification task, where the model evaluates whether a candidate label matches the input query. A logit-difference ranking scheme is proposed, providing a simple yet effective optimization objective.
    
%    \item \textbf{Knowledge distillation for efficiency:} By distilling knowledge from large language models into smaller models, the framework retains high performance while significantly reducing inference cost, making it suitable for large-scale educational applications.
%\end{itemize}

\section{Problem Formulation}
\label{sec::sys_pro}
We model misconception detection as a hierarchical multi-stage classification problem, which provides a structured formulation for capturing semantic variation in student responses.

Formally, let the dataset be defined as:
\begin{equation}
    \mathcal{D} = \{ (Q_i, A_i, E_i, y_i) \}_{i=1}^N,
\end{equation}
where each instance consists of a diagnostic multiple-choice question $Q_i$, the student’s selected answer $A_i$, the free-text explanation $E_i$, and the corresponding ground-truth label $y_i$.

Each input instance is represented as a tuple:
\begin{equation}
    x = \left( Q, A, E \right).
\end{equation}

The objective is a system $\mathcal{S}$ that maps each input to a predicted label:
\begin{equation}
    \mathcal{S}: (Q, A, E) \longrightarrow \hat{y},
\end{equation}
where $\hat{y} \in \mathcal{M}$ denotes the label predicted by the system.

Specifically, the label space $\mathcal{M}$  is structured into three hierarchical levels:
\begin{enumerate}
    \item \textbf{Answer correctness:} Determine whether the selected answer $A$ is correct or incorrect:
    \begin{equation}
        y^{(1)} \in \{ \texttt{True}, \texttt{False} \}.
    \end{equation}

    \item \textbf{Explanation quality:} Assess whether the explanation $E$ contains a misconception:
    \begin{equation}
        y^{(2)} \in \{ \texttt{Correct}, \texttt{Misconception}, \texttt{Neither} \}.
    \end{equation}

    \item \textbf{Misconception identification:} If $y^{(2)} = \texttt{Misconception}$, identify the specific misconception from a finite set $\mathcal{M'}$ of misconception labels:
    \begin{equation}
        y^{(3)} \in \mathcal{M'} \cup \{\texttt{NA}\}.
    \end{equation}
\end{enumerate}

Thus, the predicted output combines the three stages but, for misconception identification, returns a ranked list of candidate labels.

% \subsection{Problem Formulation}
% \label{sec::problem}
%Formally, let the dataset be defined as:
%\begin{equation}
%    \mathcal{D} = \{ (Q_i, A_i, E_i, y_i) \}_{i=1}^N,
%\end{equation}
%where $N$ is the number of annotated samples and $y_i$ is the ground-truth label. For each instance, the system produces a ranked list of top-$m$ candidates:
%\begin{equation}
%    \hat{Y}_i = \mathcal{S}(Q_i, A_i, E_i) = \left[ \hat{y}_{i1}, \hat{y}_{i2}, \ldots, \hat{y}_{im} \right].
%\end{equation}

%To evaluate the system, we adopt the \textbf{Mean Average Precision at $m$ (MAP@$m$)} metric. For a given instance $i$ with golden misconception label $y_i$, the score is defined as:
%\begin{equation}
%    \text{MAP@}m(i) = 
%    \begin{cases}
%    \frac{1}{j}, & \text{if } y_i = \hat{y}_{ij} \text{ for some } j \in \{1,2,\ldots,m\}, \\[6pt]
%    0, & \text{otherwise}.
%    \end{cases}
%\end{equation}

%The overall MAP@$m$ across the dataset is then given by:
%\begin{equation}
%    \text{MAP@}m(\mathcal{S}) = \frac{1}{N} \sum_{i=1}^N \text{MAP@}m(i).
%\end{equation}

%This metric rewards the system for ranking the correct misconception label higher, providing a balanced evaluation that captures both prediction accuracy and ranking quality.
\section{Motivation and Key ideas}
\subsection{Motivation}

\noindent\textit{Observation 1: Effective error detection and classification task in the education domain demands automation}

Error detection and classification are crucial in education, as they directly support learning. Consequently, developing automated systems for this task that leverage the generalization capabilities of LMs \cite{yang2025qwen3, umehara2025benchmarking} represents a promising approach to enhance both efficiency and educational outcomes.

\noindent\textit{Observation 2: Ensemble models for improved generalization
}

While many existing approaches rely on a single model to perform detection or classification tasks \cite{bewersdorff2023assessing}, such methods often face certain limitations in practice, such as noise, bias, or limited generalization. By leveraging the diversity among models, ensemble approaches \cite{fu2025rlae, fang2024llm} can mitigate individual weaknesses, reduce variance, decrease bias, and enhance the stability of predictions.

\noindent\textit{Observation 3: Retrieval and Reranking are essential to address search challenges}

Educational tasks, particularly error classification, often require navigating an ample, complex space of prior samples to detect recurring patterns in students' reasoning. Therefore, it is essential to employ systems that efficiently filter and prioritize candidate error categories, ensuring that the model attends to the most relevant possibilities. 

\noindent\textit{Observation 4: Structured reasoning supports complex decision-making}

Complex educational problems, particularly mathematics, often require multi-step reasoning that cannot be effectively addressed through direct prediction. Therefore, organizing the decision-making process into structured reasoning steps allows the model to examine candidates from multiple perspectives and make more reliable judgments.  

\subsection{Key Ideas}
\label{sec::key_ideas}
\noindent\textit{Key Idea 1: Ensemble of multiple LMs for robust decision-making}

To address the limitations of single-LM-based systems mentioned in \textit{Observation 2}, our approach incorporates an ensemble of multiple LMs, where each model independently acts as a decider for the given task. The final decision is then derived by aggregating the outputs of these individual models using specific strategies. The effectiveness of this design is empirically validated in Section~\ref{sec::MiRAGE}, where we compare single-LM and multi-LM variants of our system across different evaluation scenarios.

\noindent\textit{Key Idea 2: Efficient Candidate Selection through Retrieval and Reranking}

Building on \textit{Observation~3}, we address the challenge of navigating a large and complex space of possible error types. To manage this, our system incorporates \textbf{retrieval} \cite{lewis2020retrieval}, a process that efficiently filters and identifies a focused subset of candidate error types relevant to the student input. These candidates are subsequently refined through \textbf{reranking} \cite{huang2025gumbel}, which assigns relevance scores to prioritize the most accurate classifications. By combining retrieval and reranking, the system reduces the search space and enhances robustness and accuracy. Moreover, the evaluation of candidates can be maintained and propagated across models, supporting iterative refinement and collaborative decision-making.

\noindent\textit{Key Idea 3: Structured Reasoning for Multi-Model Collaboration}

Motivated by \textit{Observation~4}, we adopt structured, step-by-step reasoning to tackle complex educational problems. Specifically, CoT reasoning \cite{wang2024chain} is applied to enable the system to decompose each classification decision into smaller, interpretable reasoning steps, allowing the model to analyze candidates from multiple perspectives. This structured reasoning can be shared across models, serving as transferable hints rather than definitive answers, facilitating collaborative decision-making.

\section{Methodology}

\begin{figure}[t]
    \centering
    \includegraphics[width=1\linewidth]{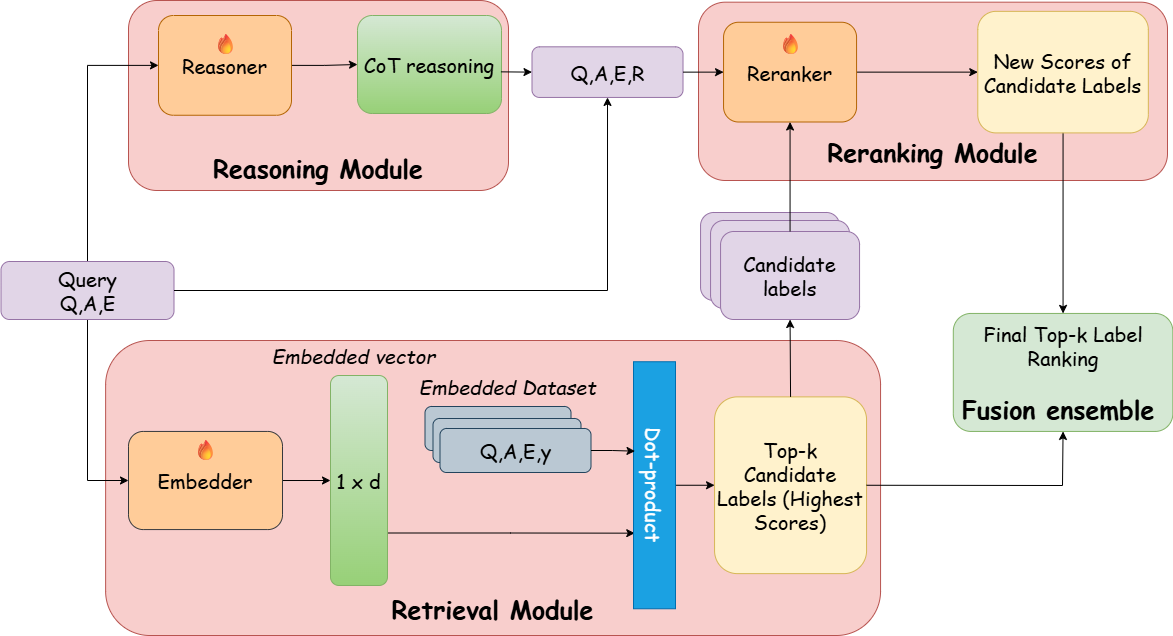}
    \caption{\textbf{The MiRAGE pipeline framework.} 
    The query is first embedded by the \textit{Retrieval} module to select top-$k$ candidate labels. In parallel, the \textit{Reasoning} module generates explanations. Both are then passed to the \textit{Reranking} module, which realigns scores with the reasoning. Finally, retrieval and reranking scores are fused through an ensemble strategy to produce the final ranking.}

    % The query is first embedded by the \textit{Embedder} in the \textit{Retrieval} module to select top-$k$ candidate labels based on semantic similarity. In parallel, the \textit{Reasoning} module provides explanatory reasoning about the query. The candidate labels and reasoning outputs are then passed to the \textit{Reranking} module to assign new scores aligned with the reasoning. Finally, scores from the \textit{Retrieval} and \textit{Reranking} modules are combined through an ensemble fusion strategy to produce the final ranking.}
    \label{fig:CIRE_qual}
\end{figure}

\subsection{Overview}
As discussed in Section~\ref{sec::key_ideas}, our framework integrates multiple modules through an ensemble mechanism to produce the final prediction. The main components are outlined as follows:

\begin{itemize}
    \item \textbf{Retrieval module:} Identifies and retrieves database samples with strong semantic similarity to the input query using an embedding model (\textit{Embedder}), enabling similarity-based predictions.

    \item \textbf{Reasoning module:} Utilizes CoT reasoning via an LM (\textit{Reasoner}) to generate structured reasoning traces, explanatory narratives, and assessments of logical inconsistencies in student responses, enhancing prediction accuracy.

    \item \textbf{Reranking module:} Employs an additional (\textit{Reranker}) to score and reorder candidates retrieved from the database, prioritizing those most consistent with the \textit{Reasoner}'s analyses, producing a refined ranking.

    \item \textbf{Fusion ensemble mechanism:} Aggregates outputs from the \textit{Retrieval} and \textit{Reranking} modules via a weighted ensemble, combining complementary strengths to yield a robust and accurate final decision.
\end{itemize} 

Together, these modules form a cohesive pipeline, as illustrated in Fig.~\ref{fig:CIRE_qual}, that integrates retrieval, reasoning, and reranking to enhance prediction accuracy and overall interpretability.

\subsection{Retrieval module}
\label{sec::retrieval}
%Our approach exploits the semantic representation of each input to retrieve samples with high semantic similarity from the dataset, thereby facilitating the generation of high-quality preliminary decisions about the underlying misconceptions.

Leveraging the strengths of embedding models in capturing semantic meaning, the \textit{Embedder} model in the \textit{Retrieval} module generates semantic representations for each triplet \((Q, A, E)\) in the dataset \(D\). This process produces an embedded dataset \(\mathcal{E}_D\), consisting of pairs \(\big(\mathbf{z}_i, y_i\big)\), where \(\mathbf{z}_i \in \mathbb{R}^d\) denotes the \(d\)-dimensional embedding of the triplet and \(y_i\) is its associated label:  
\begin{equation}
\mathcal{E}_D = \left\{ \big(\mathbf{z}_i, y_i\big) \,\middle|\, \mathbf{z}_i = f_{\text{embed}}(Q_i, A_i, E_i), \; i = 1, \dots, |D| \right\}.
\end{equation}

Given an input query triplet, represented as \(\mathbf{z}_{\text{query}}\), the similarity score between \(\mathbf{z}_{\text{query}}\) and each stored representation \(\mathbf{z}_i\) is calculated as follows:
\begin{equation}
s_i = \mathbf{z}_{\text{query}}^\top \mathbf{z}_i, \quad \forall i \in \{1, \dots, |D|\}.
\end{equation}

To aggregate similarity at the label level, the similarity score between the query and each label \(y \in \mathcal{M}\) is defined as the maximum similarity among all samples associated with that label:
\begin{equation}
\text{score}_{\text{retrieve}}(y) = \max_{\{i : y_i = y\}} s_i, \quad \forall y \in \mathcal{M}, i \in \{1, \dots, |D|\}
\end{equation}

Finally, the \textit{Retrieval} module ranks all labels in \(\mathcal{M}\) by their similarity scores \(\text{score}_{\text{retrieve}}(y)\) and selects the top-$k$ distinct labels, denoted as \(Y_k^{\text{retrieve}}\). These labels are treated as candidate misconceptions, derived from samples with strong semantic similarity to the input.

\textbf{Training the Embedder.}
Since the labels in our dataset capture inherent semantic relationships, conventional supervised contrastive learning is limited by its assumption that all classes are equally distinct \cite{khosla2020supervised}. To address this, inspired by the SupCon loss, we introduce a predefined \emph{soft similarity mask} \(M \in \mathbb{R}^{N \times N}\), where each entry \(M_{i,j}\) quantifies the relational strength between samples \(i\) and \(j\) according to their types of misconceptions. This formulation allows contrastive learning to leverage graded inter-class similarities rather than a strict binary notion of similarity.

The proposed masked supervised contrastive loss is then defined as:
\begin{equation}
\mathcal{L}_{\text{MaskSupCon}} = 
\sum_{i \in I} 
\frac{-1}{\sum_{j \in A(i)} |M_{i,j}|}
\sum_{j \in A(i)} 
M_{i,j} \cdot 
\log \frac{
\exp \left( \mathbf{z}_i \cdot \mathbf{z}_j / \tau \right)}
{\sum_{a \in A(i)} \exp \left( \mathbf{z}_i \cdot \mathbf{z}_a / \tau \right)},
\end{equation}
where $I$ is the set of all samples in a batch, $A(i)$ is the set of all candidate samples excluding $i$, $\tau$ is a temperature hyperparameter, and $\mathbf{z}_i$ is the embedding of sample $i$.  

\subsection{Reasoning module}

The \textit{Reasoner}  in our framework is instantiated as an LM that maps each input triplet to a structured CoT reasoning output, denoted as $R$. Subsequently, this CoT reasoning is forwarded to the subsequent modules, which make the final decision based on the detailed analysis provided by the \textit{Reasoner}, thereby improving both the confidence and accuracy of the outcome.

\textbf{Training the Reasoner.} 
Similar to the \textit{Embedder}, the \textit{Reasoner} in our framework is pretrained to enhance its reasoning capability in the misconception detection task in mathematical problems. Specifically, the pretraining leverages \textit{knowledge distillation}, where a teacher LLM, referred to as the \textit{CoT Teacher}, generates CoT reasoning for the corresponding inputs to guide the \textit{Reasoner}. Consequently, this approach achieves \textbf{cost efficiency}, significantly reducing the computational overhead at inference compared to directly deploying the large teacher model.

To ensure high-quality teacher rationales for distillation, we employ a two-stage \emph{generation--evaluation} pipeline that (i) generates multiple candidate CoT reasonings per example and (ii) uses an LLM judge (as motivated in \cite{kim2024biggen}) to select the best candidate. Formally, for each annotated instance
\((Q_i, A_i, E_i, y_i)\), the CoT teacher produces \(m\) reasoning candidates via stochastic sampling:
\begin{equation}
    \mathcal{C}_i = \{R_{\mathrm{teacher},i}^{(1)}, \dots, R_{\mathrm{teacher},i}^{(m)}\}.
\end{equation}

An LLM-based judge then evaluates every candidate in \(\mathcal{C}_i\) with respect to logical coherence, factual consistency, and alignment with the ground-truth label \(y_i\). The top-ranked candidate is selected:

\begin{equation}
    R_{\mathrm{teacher},i} = \arg\max_{R\in\mathcal{C}_i} \; \mathrm{Judge}(R \mid Q_i, A_i, E_i, y_i)
\end{equation}
The distilled-augmentation dataset is thus
\begin{equation}
\mathcal{R}_D = \big\{(Q_i, A_i, E_i, y_i, R_{\mathrm{teacher},i})\big\}_{i=1}^{|\mathcal{D}|}.
\end{equation}
This selective procedure filters low-quality or inconsistent rationales and provides higher-fidelity supervision for supervised fine-tuning of the \textit{Reasoner}.

The \textit{Reasoner} is then trained on \(\mathcal{R}_D\) through a \textbf{supervised fine-tuning (SFT) process}, leveraging distilled knowledge from the CoT Teacher to generate detailed reasoning chains for new inputs, thereby enhancing its ability to identify and explain potential misconceptions.

%The \textit{Reasoner} is then trained on $\mathcal{R}_D$ via a \textbf{supervised fine-tuning (SFT) process}, with the objective of minimizing the negative log-likelihood of generating the teacher reasoning given the input:

%\begin{equation}
%\mathcal{L}_{\text{SFT}} = - \sum_{i=1}^{|\mathcal{R}_D|} \log P_\theta \big(R_{\text{teacher}, i} \mid Q_i, A_i, E_i, y_i \big),
%\end{equation}

%where $\theta$ denotes the parameters of the \textit{Reasoner}.

%Through this process, the \textit{Reasoner} learns to generate detailed reasoning chains for new inputs, enabling it to identify and explain potential misconceptions more effectively.

\subsection{Reranking Module} 

%The \textit{Reranking} module functions in close collaboration with both the \textit{Retrieval} and \textit{Reasoning} modules. Its primary objective is to refine the set of candidate labels generated by the \textit{Retrieval} module by incorporating the detailed reasoning provided by the \textit{Reasoning} module for the mathematical input $(Q, A, E)$. In particular, the \textit{Reranking} module assesses the consistency of each candidate label with the input query, conditioned on the reasoning $R$, thereby producing a more reliable and accurate ranking of potential misconceptions.

Formally, the \textit{Reranker} model in the \textit{Reranking} module takes as input the query $(Q, A, E)$, the corresponding reasoning $R$ from the \textit{Reasoning} module, and the set of $k$ candidate labels $Y_k^{\text{retrieve}}$ generated by the \textit{Retrieval} module. For each candidate label $y \in Y_k^{\text{retrieve}}$, a \textbf{prompt}, as shown in Figure~\ref{prompt_rerank}, is constructed by combining $(Q, A, E)$, $R$, and $y$. The model is instructed to output a single token: "Yes" if the label is consistent with both the query and the reasoning $R$, and "No" otherwise. This interation is formalized as:

\begin{equation}
    \text{Reranker}\big((Q,A,E),R,y\big) \;\rightarrow\; 
\begin{cases}
\text{Yes}, & \text{if $y$ matches}, \\
\text{No}, & \text{otherwise}.
\end{cases}
\end{equation}
\begin{figure}[htbp]
\centering
\begin{tikzpicture}[x=1cm,y=1cm]
\def\W{13}
\def\H{21}
\def\HeaderH{1}

\draw (0,0) rectangle (\W,\H);

\fill[gray!40] (0,\H) rectangle (\W,\H-\HeaderH);
\node[anchor=west] at (0.3,\H-0.65) {\textbf{Re-ranker module 's input}};
\fill[gray!10] (0,0) rectangle (\W,\H-\HeaderH);
\node[anchor=west,text width=12cm] at (0.3,\H-1.3) {\texttt{<|im\_start|> system}};
\node[anchor=west, text width=12cm] at (0.3,\H-2.8) {\texttt{
You are a meticulous educational analyst and expert in mathematics pedagogy. Your task is to perform a verification check. You will be given a student's response to a math problem, then a THOUGHT ANALYSIS and a proposed classification for that response. You must determine if the proposed classification is entirely accurate based on your knowledge and problem data.
}};

% Bỏ dòng trắng
\node[anchor=west] at (0.3,\H-4.3) {\textbf{DEFINITIONS OF THE CLASSIFICATION LABELS:}};

\node[anchor=west,text=blue] at (0.3,\H-4.9) {\textbf{Part 1: Correctness (True or False).}};
\node[anchor=west,] at (0.3,\H-5.3) {
\texttt{This describes whether the student's answer is objectively correct.}
};
\node[anchor=west,text=blue] at (0.3,\H-5.7) {\textbf{Part 2: ReasoningType (Correct, Misconception, or Neither).}};
\node[anchor=west,text width = 12cm] at (0.3,\H- 6.3) {\textit{Correct}\texttt{: The explanation shows sound, logical, and mathematically valid reasoning.}};
\node[anchor=west,text width=12cm] at (0.3,\H-7.1) {
\textit{Misconception}\texttt{: The explanation reveals a specific, identifiable error in conceptual understanding.}};
\node[anchor=west,text width=12cm] at (0.3,\H-8.1) {
\textit{Neither}\texttt{: The explanation is incorrect, but does not point to a specific misconception. If the ReasoningType is Correct or Neither, the Misconception field should be "NA".}};
\node[anchor = west,text = blue] at (0.3, \H-8.9){
\textbf{Part 3: Misconception (Misconception type)}
};
\node[anchor = west,text width = 12cm] at (0.3, \H- 9.9){
\texttt{This is a text description of the specific thinking error. It is only relevant when the ReasoningType is Misconception. If the ReasoningType is Correct or Neither, this field's value should be "NA".}
};
\node[anchor=west] at (0.3,\H-11.0) {\textbf{YOUR TASK:}};
\node[anchor=west,text width=12cm] at (0.3,\H-12.1) {
\texttt{Compare the } \textbf{THOUGHT ANALYSIS} \texttt{ to the Correctness, ReasoningType and Misconception in }\textbf{PROPOSED CLASSIFICATION}\texttt{. Then,  output "Yes", if you think the \textbf{PROPOSED CLASSIFICAION} is correct. Else, output "No".}};
\node[anchor=west] at (0.3, \H - 13.2){
\textbf{CONSTRAINT:}
};
\node[anchor=west] at (0.3, \H - 13.65){
\texttt{You are only allowed to output only one token ("Yes"/"No").}
};
\node[anchor=west] at (0.3,\H-14.1) {\texttt{<|im\_end|>}}; 
\node[anchor=west] at (0.3,\H-14.6) {\texttt{<|im\_start|>user}}; 
\node[anchor=west] at (0.3,\H-15.1) {\textbf{PROBLEM DATA:}}; 
\node[anchor=west] at (0.3,\H-15.6) {\texttt{Question: \{q\_text\}}};
\node[anchor=west] at (0.3,\H-16.1) {\texttt{Student's Answer: \{mc\_answer\}}};
\node[anchor=west] at (0.3,\H-16.6) {\texttt{Student's Explanation: \{explanation\}}};
\node[anchor=west] at (0.3,\H-17.1) {\textbf{PROPOSED CLASSIFICATION:}}; 
\node[anchor=west] at (0.3,\H-17.6) {\texttt{Correctness: \{correctness\}}};
\node[anchor=west] at (0.3,\H-18.1) {\texttt{ReasoningType: \{reasoning\_type\}}};
\node[anchor=west] at (0.3,\H-18.6) {\texttt{Misconception: \{misconception\}}};
\node[anchor=west] at (0.3,\H-19.1) {\textbf{THOUGHT ANALYSIS:}}; 
\node[anchor=west] at (0.3,\H-19.6) {\texttt{\{thought\}}};
\node[anchor=west] at (0.3,\H-20.1) {\texttt{<|im\_end|>}};
\node[anchor=west] at (0.3,\H-20.6) {\texttt{<|im\_start|>assistant}};
\end{tikzpicture}
\caption{Prompt for re-ranker module}
\label{prompt_rerank}
\end{figure}
The candidate labels are sequentially reranked based on the confidence of the \textit{Reranker}’s decisions. Specifically, let $\ell_{\text{Yes}}(y)$ and $\ell_{\text{No}}(y)$ denote the logits assigned by the \textit{Reranker} to the output tokens ``Yes'' and ``No'', respectively. The reranking score for each label $y$ is then computed as the \textbf{logit margin}, where a larger margin indicates stronger agreement between the label $y$ and the input $(Q, A, E)$ along with the associated reasoning $R$:

\begin{equation}
\text{score}_{\text{rerank}}(y) = \ell_{\text{Yes}}(y) - \ell_{\text{No}}(y), 
\quad \forall y \in Y_k^{\text{retrieve}}.
\end{equation} 

\textbf{Training the Reranker.}
In our framework, the \textit{Reranker} is trained to generate accurate predictions by adhering to the structured format when given a corresponding prompt and its reasoning. To achieve this, we exploit the strength of a \textbf{Data Augmentation} process. Specifically, from the dataset with reasoning \(\mathcal{R}_D := (Q, A, E, R,y)\), previously defined, we use the actual samples as positive instances. In particular, for these positive instances, the \textit{Reranker} is trained to learn the mapping.
\begin{equation}
    \text{Prompt}(Q,A,E,R,y) \mapsto \text{Yes}.
\end{equation}

Meanwhile, for each true sample \((Q_i, A_i, E_i, R_i, y_i)\), we construct a set of \(m\) negative samples by randomly selecting \(m\) labels \(\{\tilde{y}_{i1}, \dots, \tilde{y}_{im}\}\) such that \(\tilde{y}_{ij} \neq y_i\). These negative instances are expressed as:
\[
\text{Prompt}(Q_i, A_i, E_i, R_i, \tilde{y}_{ij}) \mapsto \text{No}, \quad j=1,\dots,m.
\]

By collecting both positive and negative samples, a new training dataset is constructed for the \textit{Reranker}. The model is then optimized using a \textbf{Cross-Entropy loss} objective:
\begin{equation}
\mathcal{L}_{\text{rerank}} = - \frac{1}{N} \sum_{i=1}^{|\mathcal{R}_D|} \sum_{j=1}^{m+1} \Big[ y_{ij} \log p_{ij} + (1-y_{ij}) \log (1-p_{ij}) \Big],
\end{equation}
where \(p_{ij}\) is the predicted probability assigned by the Reranker to the \(j\)-th candidate of the \(i\)-th query, and \(y_{ij} \in \{0,1\}\) denotes whether it is a true or negative sample.

\subsection{Fusion Ensemble Mechanism}
To enhance prediction robustness, our framework employs an ensemble strategy that combines the complementary strengths of the \textit{Retrieval} and \textit{Reranking} modules. Specifically, to obtain the final decision for a given query, we compute a \textbf{weighted fusion} of the scores assigned by both modules. Since the raw scores of the two modules may lie on different scales, we first normalize them individually using the \textbf{softmax} function to become comparable. After this normalization step, the fusion score is computed as:
\begin{equation}
\text{score}_{\text{total}}(y) = \alpha \cdot \text{score}_{\text{rerank}}(y) + \beta \cdot \text{score}_{\text{retrieve}}(y),
\end{equation}
where $\alpha$ and $\beta$ are tunable hyperparameters that balance the relative contributions of the two modules. 

This fusion mechanism integrates the reasoning-aware evaluation of the \textit{Reranker} with the broader coverage of the \textit{Retriever}. By combining these complementary perspectives, the framework benefits from the fine-grained semantic alignment offered by the \textit{Reranker} and preserves the diversity and inclusiveness ensured by the \textit{Retriever}. This dual contribution helps balance precision and recall, preventing the system from being overly narrow in its predictions while maintaining interpretability and robustness. 
\section{Empirical Evaluation}
\label{sec:experiments}

We evaluate the effectiveness of our proposed framework by addressing two main research questions:

{\hangindent=2em \textbf{RQ1. [Ensemble Model]} How does the \textbf{MiRAGE} model compare against using individual modules in isolation?}

{\hangindent=2em \textbf{RQ2. [Ablation Analysis]} What extent does each component of our method contribute to overall performance?}

All experiments were conducted on a system equipped with an NVIDIA A100 GPU (80GB PCIe).

{\hangindent=2em \textbf{Dataset.} We conduct experiments on the MAP Student Misconceptions dataset provided on Kaggle\footnote{\url{https://www.kaggle.com/competitions/map-charting-student-math-misunderstandings/data}}, which was released as part of the Mathematical Assessment of Performance (MAP) competition. This dataset contains student responses to mathematics problems, annotated with fine-grained misconception labels.}

{\hangindent=2em \textbf{Metrics.} System performance is evaluated using the \textbf{Mean Average Precision at $m$ (MAP@$m$)} metric, which measures both accuracy and ranking quality. For an instance $i$ with ground-truth label $y_i$, the score is:
\begin{equation}
    \text{MAP@}m(i) = 
    \begin{cases}
    \tfrac{1}{j}, & \text{if } y_i = \hat{y}_{ij} \text{ for some } j \in \{1,2,\ldots,m\}, \\[6pt]
    0, & \text{otherwise},
    \end{cases}
\end{equation}
where $\hat{y}_{ij}$ denotes the $j$-th ranked prediction for instance $i$. The overall MAP@$m$ is then computed as:
\begin{equation}
    \text{MAP@}m(\mathcal{S}) = \frac{1}{N} \sum_{i=1}^N \text{MAP@}m(i),
\end{equation}
with $N$ being the total number of instances. In our experiments, we report results on MAP@1, MAP@3, and MAP@5.}

{\hangindent=2em \textbf{Model.} In the \textit{Retrieval} module, we employ \textbf{MathBERT} as the \textit{Embedder} to capture semantic representations of student answers. The \textit{Reasoning} module is built on the \textbf{Qwen3-8B} \textit{Reasoner}, further enhanced through distillation from the \textit{CoT Teacher} \textbf{GPT-OSS-20B}. For the \textit{Reranking} stage, we adopt \textbf{Qwen3-7B} as the \textit{Reranker}.}

%{\hangindent=2em \textbf{Similarity Mask.} Given that each label is represented as a triplet $y = [y^{(1)}, y^{(2)}, y^{(3)}]$, the \textit{Similarity Mask} $M$ for \textit{Training the Embedder}, as defined in Section \ref{sec::retrieval}, is constructed as follows:}
%\begin{equation}
%M_{i,j} =
%\begin{cases}
%1.0, & \text{if } (y_i^{(1)} = y_j^{(1)}) \land (y_i^{(2)} = y_j^{(2)}) \land (y_i^{(3)} = y_j^{(3)}), \\[6pt]
%0.7, & \text{if } (y_i^{(1)} = y_j^{(1)}) \land (y_i^{(2)} = y_j^{(2)}) \land (y_i^{(3)} \neq y_j^{(3)}), \\[6pt]
%0.4, & \text{if } (y_i^{(1)} = y_j^{(1)}) \land (y_i^{(2)} \neq y_j^{(2)}), \\[6pt]
%$0.3, & \text{if } (y_i^{(1)} \neq y_j^{(1)}) %\land (y_i^{(2)} = y_j^{(2)}) \land (y_i^{(3)} = $y_j^{(3)}), \\[6pt]
%0.1, & \text{if } (y_i^{(1)} \neq y_j^{(1)}) \land (y_i^{(2)} = y_j^{(2)}) \land (y_i^{(3)} \neq y_j^{(3)}), \\[6pt]
%0.0, & \text{otherwise}.
%\end{cases}
%\end{equation}

{\hangindent=2em \textbf{Optimizer.} All models are fine-tuned using \textbf{LoRA} \cite{hu2021loralowrankadaptationlarge}. The ensemble mechanism combines module outputs through weighted fusion, with coefficients $\alpha = 0.7$ and $\beta = 0.3$ controlling the relative contributions of the \textit{Reranker} and \textit{Retriever}.}

\section{Performance of MiRAGE}
\label{sec::MiRAGE}
To assess the effectiveness of our approach, we conducted experiments on the dataset, comparing the proposed \textsc{MiRAGE} system against several baselines. In particular, to evaluate the impact of the \textit{ensemble strategy} in \textsc{MiRAGE}, the baselines were constructed using the outputs of the \textit{Reranking Module} and the \textit{Retrieval Module} individually, while \textsc{MiRAGE} combines them through \textit{Fusion Ensemble Mechanism}. As shown in Table~\ref{tab:rq1}, \textsc{MiRAGE} consistently outperforms all baselines by a significant margin.

% TABLE I
\begin{table}[H]
\small
\centering
\caption{Performance comparison of \textsc{MiRAGE} with individual modules}\label{tab:rq1}
\begin{tabularx}{\columnwidth}{l *{4}{>{\centering\arraybackslash}X}}
\toprule
Method & MAP@1-Score & MAP@3-Score & MAP@5-Score  \\
\midrule
Reranking Module & 0.79 & 0.81 & 0.88 \\
Retrieval Module & 0.74 & 0.83 & 0.85 \\
\textbf{MiRAGE (essemble)} & \textbf{0.82} & \textbf{0.92} & \textbf{0.93} \\
\bottomrule
\end{tabularx}
\end{table}
The effectiveness of \textsc{MiRAGE} consistently surpasses its individual modules across all metrics (MAP@1, MAP@3, MAP@5). For instance, under MAP@1, the \textit{Reranking Module} scores 0.79, while the \textit{Retrieval Module} reaches only 0.74. The weaker performance of the \textit{Retrieval Module} arises from its reliance on semantic similarity without the capacity for multi-faceted reasoning. In contrast, the \textit{Reranking Module} remains vulnerable to biases from both the \textit{Reasoner} and the \textit{Reranker}, which, despite their detailed analyses, often struggle to distinguish between different misconceptions that are semantically similar. By integrating these complementary strengths, \textsc{MiRAGE} achieves 0.82 on MAP@1.

On other metrics, \textit{Reranking} and \textit{Retrieval} obtain 0.81 vs. 0.83 on MAP@3 and 0.88 vs. 0.85 on MAP@5, whereas \textsc{MiRAGE} achieves 0.92 and 0.93, respectively. These results confirm that ensembling mitigates individual weaknesses: \textit{Retrieval} ensures broad coverage but is less precise at top ranks. In contrast, \textit{Reranking} offers higher accuracy but relies on initial retrieval quality. Integrated, the ensemble boosts MAP@1 and maintains consistent gains on MAP@3 and MAP@5, enhancing accuracy and coverage.

\section{Effect of Individual Components}
In this section, we examine the performance of the full \textsc{MiRAGE} model against its ablated variants to assess the contribution of individual components. Specifically, we consider two reduced configurations: (i) \textbf{Without Fine-tuned Reasoner}, where the \textit{Reasoner} within the \textit{Reranking} module relies solely on the pretrained backbone without task-specific adaptation, and (ii) \textbf{Without Fine-tuned Reranker}, where the \textit{Reranker} is not fine-tuned to reorder candidate labels based on the given problems and the corresponding reasoning provided by the \textit{Reasoner}.

 % TABLE III
\begin{table}[H]
\small
\centering
\caption{Component-wise Evaluation of \textsc{MiRAGE} Across Metrics}
\label{tab:ablation-FLUID}
\begin{tabularx}{\columnwidth}{l *{4}{>{\centering\arraybackslash}X}}
\toprule
Method & MAP@1-Score & MAP@3-Score  & MAP@5-Score \\
\midrule
Without Fine-tuned Reasoner & 0.54 \textcolor{red}{{(-0.28)}} & 0.60 \textcolor{red}{{(-0.32)}} & 0.66 \textcolor{red}{{(-0.27)}}\\
Without Fine-tuned Reranker & 0.63 \textcolor{red}{{(-0.19)}} & 0.75 \textcolor{red}{{(-0.17)}}& 0.76\textcolor{red}{{(-0.17)}} \\
\textbf{MiRAGE} & \textbf{0.82} & \textbf{0.92} & \textbf{0.93} \\
\bottomrule
\end{tabularx}
\end{table}

\textbf{Fine-tuned Reasoner:} Table~\ref{tab:ablation-FLUID} shows that removing the distillation-based fine-tuning of the \textit{Reasoner} model leads to a substantial performance decline: MAP@1 drops to 0.54, MAP@3 to 0.60, and MAP@5 to 0.66. Without task-specific adaptation, the pretrained \textit{Reasoner} fails to capture subtle patterns in students’ erroneous logical reasoning, leading to a significant degradation in performance. By contrast, distillation-based fine-tuning from a \textit{CoT Teacher} LLM equips the \textit{Reasoner} with domain-specific reasoning abilities while eliminating the need for large-scale LLMs in the inference pipeline. 

%\textbf{Fine-tuned Reasoner:} Table~\ref{tab:ablation-FLUID} shows that removing the distillation-based fine-tuning of the \textit{Reasoner} model leads to a substantial performance decline: MAP@1 drops to 0.54, MAP@3 to 0.60, and MAP@5 to 0.66. These results emphasize the critical role of fine-tuning in strengthening the \textit{Reasoner}' effectiveness, as well as the overall performance of the MiRAGE system. Without task-specific adaptation, the pretrained \textit{Reasoner} fails to capture subtle patterns in students’ erroneous logical reasoning, leading to a significant degradation in performance. By contrast, distillation-based fine-tuning from a \textit{CoT Teacher} LLM equips the \textit{Reasoner} with domain-specific reasoning abilities while eliminating the need for large-scale LLMs in the inference pipeline. This demonstrates both the effectiveness of the proposed knowledge distillation process and the capability of small-to-medium LMs when properly fine-tuned.

\textbf{Fine-tuned Reranker:}  
Similarly, removing the fine-tuning of the \textit{Reranker} model results in a substantial performance decline: a 23\% drop in MAP@1, 18\% in MAP@3, and 18\% in MAP@5. Although the \textit{Reasoner} provides strong reasoning and detailed analysis of problems, the task of identifying the correct label based on this information remains highly challenging, as the dataset contains many labels with high semantic similarity. Exploiting a data augmentation strategy, the proposed fine-tuning process enables the model not only to align problems and their reasoning with the correct label but also to discriminate effectively against closely related alternatives, thereby boosting the overall performance of the \textsc{MiRAGE} system.

\textbf{Summary.} The results highlights the critical importance of both the \textit{Reasoner} and the \textit{Reranker} within the \textsc{MiRAGE} architecture. Their interdependence ensures that the ensemble effectively integrates semantic coverage with fine-grained reasoning, ultimately driving the superior performance of \textsc{MiRAGE}.

\section{Conclusion and Future works}
In this study, we present \textsc{MiRAGE}, a cost-effective hybrid framework for misconception detection in mathematics. The framework integrates three key components into a cohesive pipeline: a retrieval module that leverages semantic similarity for candidate selection, a reasoning module that applies CoT prompting to capture logical inconsistencies, and a reranking module that refines results through cross-attention scoring. Their outputs are combined via a fusion ensemble mechanism to yield robust predictions. By exploiting the complementary strengths of each module, \textsc{MiRAGE} achieves a balance of interpretability, accuracy, and efficiency. Experimental results demonstrate that \textsc{MiRAGE} consistently outperforms baselines that rely on single-module decisions on MAP score metrics, validating the effectiveness of the ensemble strategy. Moreover, additional findings confirm that each module's proposed training and fine-tuning strategies are crucial for maximizing performance. These results showcase \textsc{MiRAGE}'s ability to detect misconceptions at scale while maintaining lower computational costs than LLMs.

For future work, we aim to extend \textsc{MiRAGE} to domains beyond mathematics, including science and language learning, where misconceptions are equally important. We also plan to explore the integration of multimodal data, such as diagrams and handwritten solutions, to better capture student reasoning and enhance detection accuracy.
 
\bibliographystyle{IEEEtran}
\bibliography{citation}

\begin{thebibliography}{10}
\providecommand{\url}[1]{#1}
\csname url@samestyle\endcsname
\providecommand{\newblock}{\relax}
\providecommand{\bibinfo}[2]{#2}
\providecommand{\BIBentrySTDinterwordspacing}{\spaceskip=0pt\relax}
\providecommand{\BIBentryALTinterwordstretchfactor}{4}
\providecommand{\BIBentryALTinterwordspacing}{\spaceskip=\fontdimen2\font plus
\BIBentryALTinterwordstretchfactor\fontdimen3\font minus \fontdimen4\font\relax}
\providecommand{\BIBforeignlanguage}[2]{{%
\expandafter\ifx\csname l@#1\endcsname\relax
\typeout{** WARNING: IEEEtran.bst: No hyphenation pattern has been}%
\typeout{** loaded for the language `#1'. Using the pattern for}%
\typeout{** the default language instead.}%
\else
\language=\csname l@#1\endcsname
\fi
#2}}
\providecommand{\BIBdecl}{\relax}
\BIBdecl

\bibitem{fanni2023natural}
S.~C. Fanni, M.~Febi, G.~Aghakhanyan, and E.~Neri, ``Natural language processing,'' in \emph{Introduction to artificial intelligence}.\hskip 1em plus 0.5em minus 0.4em\relax Springer, 2023, pp. 87--99.

\bibitem{zhao2023survey}
W.~X. Zhao, K.~Zhou, J.~Li, T.~Tang, X.~Wang, Y.~Hou, Y.~Min, B.~Zhang, J.~Zhang, Z.~Dong \emph{et~al.}, ``A survey of large language models,'' \emph{arXiv preprint arXiv:2303.18223}, vol.~1, no.~2, 2023.

\bibitem{achiam2023gpt}
J.~Achiam, S.~Adler, S.~Agarwal, L.~Ahmad, I.~Akkaya, F.~L. Aleman, D.~Almeida, J.~Altenschmidt, S.~Altman, S.~Anadkat \emph{et~al.}, ``Gpt-4 technical report,'' \emph{arXiv preprint arXiv:2303.08774}, 2023.

\bibitem{yang2025qwen3}
A.~Yang, A.~Li, B.~Yang, B.~Zhang, B.~Hui, B.~Zheng, B.~Yu, C.~Gao, C.~Huang, C.~Lv \emph{et~al.}, ``Qwen3 technical report,'' \emph{arXiv preprint arXiv:2505.09388}, 2025.

\bibitem{team2025gemma}
G.~Team, A.~Kamath, J.~Ferret, S.~Pathak, N.~Vieillard, R.~Merhej, S.~Perrin, T.~Matejovicova, A.~Ram{\'e}, M.~Rivi{\`e}re \emph{et~al.}, ``Gemma 3 technical report,'' \emph{arXiv preprint arXiv:2503.19786}, 2025.

\bibitem{secolsky2016thinkaloud}
\BIBentryALTinterwordspacing
C.~Secolsky, T.~P. Judd, E.~Magaram, S.~H. Levy, B.~Kossar, and G.~Reese, ``Using think-aloud protocols to uncover misconceptions and improve developmental math instruction: An exploratory study,'' \emph{Numeracy}, vol.~9, no.~1, p. Article 6, 2016. [Online]. Available: \url{https://digitalcommons.usf.edu/numeracy/vol9/iss1/art6}
\BIBentrySTDinterwordspacing

\bibitem{michalenko2017dataminingtextualresponsesuncover}
\BIBentryALTinterwordspacing
J.~J. Michalenko, A.~S. Lan, and R.~G. Baraniuk, ``Data-mining textual responses to uncover misconception patterns,'' 2017. [Online]. Available: \url{https://arxiv.org/abs/1703.08544}
\BIBentrySTDinterwordspacing

\bibitem{wei2023chainofthoughtpromptingelicitsreasoning}
\BIBentryALTinterwordspacing
J.~Wei, X.~Wang, D.~Schuurmans, M.~Bosma, B.~Ichter, F.~Xia, E.~Chi, Q.~Le, and D.~Zhou, ``Chain-of-thought prompting elicits reasoning in large language models,'' 2023. [Online]. Available: \url{https://arxiv.org/abs/2201.11903}
\BIBentrySTDinterwordspacing

\bibitem{kostina2025llmTextClassification}
\BIBentryALTinterwordspacing
A.~Kostina, M.~D. Dikaiakos, D.~Stefanidis, and G.~Pallis, ``Large language models for text classification: Case study and comprehensive review,'' 2025. [Online]. Available: \url{https://arxiv.org/abs/2501.08457}
\BIBentrySTDinterwordspacing

\bibitem{setiawan2024reranking}
\BIBentryALTinterwordspacing
H.~Setiawan, ``Accurate knowledge distillation with n-best reranking,'' 2024. [Online]. Available: \url{https://arxiv.org/abs/2305.12057}
\BIBentrySTDinterwordspacing

\bibitem{calderon2023systematicstudyknowledgedistillation}
\BIBentryALTinterwordspacing
N.~Calderon, S.~Mukherjee, R.~Reichart, and A.~Kantor, ``A systematic study of knowledge distillation for natural language generation with pseudo-target training,'' 2023. [Online]. Available: \url{https://arxiv.org/abs/2305.02031}
\BIBentrySTDinterwordspacing

\bibitem{huang2024improvingacademicskillsassessment}
\BIBentryALTinterwordspacing
X.~Huang, Y.~Wu, D.~Zhang, J.~Hu, and Y.~Long, ``Improving academic skills assessment with nlp and ensemble learning,'' 2024. [Online]. Available: \url{https://arxiv.org/abs/2409.19013}
\BIBentrySTDinterwordspacing

\bibitem{jia2024reviewhybridensembledeep}
\BIBentryALTinterwordspacing
J.~Jia, W.~Liang, and Y.~Liang, ``A review of hybrid and ensemble in deep learning for natural language processing,'' 2024. [Online]. Available: \url{https://arxiv.org/abs/2312.05589}
\BIBentrySTDinterwordspacing

\bibitem{Ganaie_2022}
\BIBentryALTinterwordspacing
M.~Ganaie, M.~Hu, A.~Malik, M.~Tanveer, and P.~Suganthan, ``Ensemble deep learning: A review,'' \emph{Engineering Applications of Artificial Intelligence}, vol. 115, p. 105151, Oct. 2022. [Online]. Available: \url{http://dx.doi.org/10.1016/j.engappai.2022.105151}
\BIBentrySTDinterwordspacing

\bibitem{umehara2025benchmarking}
K.~Umehara, J.~Ota, T.~Nishii, R.~Kishimoto, and T.~Ishida, ``Benchmarking gpt-5 performance and repeatability on the japanese national examination for radiological technologists over the past decade (2016--2025),'' \emph{medRxiv}, pp. 2025--08, 2025.

\bibitem{bewersdorff2023assessing}
A.~Bewersdorff, K.~Se{\ss}ler, A.~Baur, E.~Kasneci, and C.~Nerdel, ``Assessing student errors in experimentation using artificial intelligence and large language models: A comparative study with human raters,'' \emph{Computers and Education: Artificial Intelligence}, vol.~5, p. 100177, 2023.

\bibitem{fu2025rlae}
Y.~Fu, Y.~Zhu, J.~Chai, G.~Yin, W.~Lin, Q.~Zhang, and D.~Zhao, ``Rlae: Reinforcement learning-assisted ensemble for llms,'' \emph{arXiv preprint arXiv:2506.00439}, 2025.

\bibitem{fang2024llm}
C.~Fang, X.~Li, Z.~Fan, J.~Xu, K.~Nag, E.~Korpeoglu, S.~Kumar, and K.~Achan, ``Llm-ensemble: Optimal large language model ensemble method for e-commerce product attribute value extraction,'' in \emph{Proceedings of the 47th International ACM SIGIR Conference on Research and Development in Information Retrieval}, 2024, pp. 2910--2914.

\bibitem{lewis2020retrieval}
P.~Lewis, E.~Perez, A.~Piktus, F.~Petroni, V.~Karpukhin, N.~Goyal, H.~K{\"u}ttler, M.~Lewis, W.-t. Yih, T.~Rockt{\"a}schel \emph{et~al.}, ``Retrieval-augmented generation for knowledge-intensive nlp tasks,'' \emph{Advances in neural information processing systems}, vol.~33, pp. 9459--9474, 2020.

\bibitem{huang2025gumbel}
S.~Huang, Z.~Ma, J.~Du, C.~Meng, W.~Wang, J.~Leng, M.~Guo, and Z.~Lin, ``Gumbel reranking: Differentiable end-to-end reranker optimization,'' \emph{arXiv preprint arXiv:2502.11116}, 2025.

\bibitem{wang2024chain}
X.~Wang and D.~Zhou, ``Chain-of-thought reasoning without prompting,'' \emph{Advances in Neural Information Processing Systems}, vol.~37, pp. 66\,383--66\,409, 2024.

\bibitem{khosla2020supervised}
P.~Khosla, P.~Teterwak, C.~Wang, A.~Sarna, Y.~Tian, P.~Isola, A.~Maschinot, C.~Liu, and D.~Krishnan, ``Supervised contrastive learning,'' \emph{Advances in neural information processing systems}, vol.~33, pp. 18\,661--18\,673, 2020.

\bibitem{kim2024biggen}
S.~Kim, J.~Suk, J.~Y. Cho, S.~Longpre, C.~Kim, D.~Yoon, G.~Son, Y.~Cho, S.~Shafayat, J.~Baek \emph{et~al.}, ``The biggen bench: A principled benchmark for fine-grained evaluation of language models with language models,'' \emph{arXiv preprint arXiv:2406.05761}, 2024.

\bibitem{hu2021loralowrankadaptationlarge}
\BIBentryALTinterwordspacing
E.~J. Hu, Y.~Shen, P.~Wallis, Z.~Allen-Zhu, Y.~Li, S.~Wang, L.~Wang, and W.~Chen, ``Lora: Low-rank adaptation of large language models,'' 2021. [Online]. Available: \url{https://arxiv.org/abs/2106.09685}
\BIBentrySTDinterwordspacing

\end{thebibliography}

% \printbibliography

\end{document}